\documentclass[11pt,a4paper]{article}
\usepackage[hyperref]{emnlp2018}
\usepackage{times}
\usepackage{graphicx}
\usepackage{amsmath,amssymb}
\usepackage{latexsym}
\usepackage{multirow}
\usepackage{subfig}
\usepackage{paralist}
\usepackage{url}
\usepackage{booktabs}
\usepackage{bm}

\aclfinalcopy 

\title{On Tree-Based Neural Sentence Modeling}

\author{Haoyue Shi $^{\dagger, \ddagger, }$\thanks{~Now at Toyota Technological Institute at Chicago, {\tt freda@ttic.edu}. This work was done when HS was an intern researcher at ByteDance AI Lab. } ~~~~~~ Hao Zhou$^{\ddagger}$ ~~~~~~ Jiaze Chen$^{\ddagger}$ ~~~~~~ Lei Li$^{ \ddagger}$ \\
$\dagger$: School of EECS, Peking University, Beijing, China \\
{\tt hyshi@pku.edu.cn}
\\
$\ddagger$: ByteDance AI Lab, Beijing, China \\
{\tt \{zhouhao.nlp, chenjiaze, lileilab\}@bytedance.com}}

\date{}

\begin{document}
\maketitle
\begin{abstract}
Neural networks with tree-based sentence encoders have shown better results on many downstream tasks.
Most of existing tree-based encoders adopt 
syntactic parsing trees as the explicit structure prior. 
To study the effectiveness of different tree structures, we replace the parsing trees with trivial trees (\textit{i.e.}, binary balanced tree, left-branching tree and right-branching tree) in the encoders. 
Though trivial trees contain no syntactic information, 
those encoders get competitive or even better results on all of the ten downstream tasks we investigated. 
This surprising result indicates that explicit syntax guidance may not be the main contributor to the superior performances of tree-based neural sentence modeling.
Further analysis show that tree modeling gives better results when crucial words are closer to the final representation.
Additional experiments give more clues on how to design an effective tree-based encoder.
Our code is open-source and available at \url{https://github.com/ExplorerFreda/TreeEnc}. 
\end{abstract}

\section{Introduction}

Sentence modeling is a crucial problem in natural language processing~(NLP).
Recurrent neural networks with long short term memory \cite{hochreiter1997long} or gated recurrent units  \cite{cho2014learning}
are commonly used sentence modeling approaches. 
These models embed sentences into a vector space and the resulting vectors can be used for classification or sequence generation in the downstream tasks.

In addition to the plain sequence of hidden units, recent work on sequence modeling proposes to impose tree structure 
in the encoder \cite{socher2013recursive,tai2015improved,zhu2015long}. 
These tree-based LSTMs introduce syntax tree as an intuitive structure prior for sentence modeling.
They have already obtained promising results in many NLP tasks, such as natural language inference \cite{bowman2016fast,chen2017enhanced} and machine translation \cite{eriguchi2016tree,chen2017improved,chen2017neural,P17-2092}.
\newcite{li2015tree} empirically concludes that syntax tree-based sentence modeling are effective for tasks requiring relative long-term context features.

On the other hand, some works propose to abandon the syntax tree 
but to adopt the latent tree for sentence modeling \cite{choi2017learning,yogatama2017learning,maillard2017jointly,williams2018latent}. 
Such latent trees are directly learned from the downstream task with reinforcement learning \cite{williams1992simple} or Gumbel Softmax \cite{jang2016categorical,maddison2016concrete}.
However, \newcite{williams2018latent} empirically show that, Gumbel softmax produces unstable latent trees with the same hyper-parameters but different initializations, while reinforcement learning \cite{williams2018latent} even tends to generate left-branching trees.
Neither gives meaningful latent trees in syntax, but each method still obtains considerable improvements in performance.
This indicates that syntax may not be the main contributor to the performance gains.

With the above observation, we bring up the following questions: 
What does matter in tree-based sentence modeling?
If tree structures are necessary in encoding the sentences, what mostly contributes to the improvement in downstream tasks?
We attempt to investigate the driving force of the improvement by latent trees without syntax. 

In this paper, we empirically study the effectiveness of tree structures in sentence modeling. 
We compare the performance of bi-LSTM and five tree LSTM encoders with different tree layouts, including the syntax tree, latent tree~(from Gumbel softmax) and three kinds of designed trivial trees~(binary balance tree, left-branching tree and right-branching tree).
Experiments are conducted on 10 different tasks, which are grouped into three categories, namely the single sentence classification (5 tasks), sentence relation classification (2 tasks), and sentence generation (3 tasks).
These tasks depend on different granularities of features, and the comparison among them can help us learn more about the results.
We repeat all the experiments 5 times and take the average to avoid the instability caused by random initialization of deep learning models.

We get the following conclusions:
\begin{compactitem}
\item Tree structures are helpful to sentence modeling on classification tasks, especially for tasks which need global~(long-term) context features, which is consistent with previous findings \cite{li2015tree}.
\item Trivial trees outperform syntactic trees, indicating that syntax may not be the main contributor to the gains of tree encoding, at least on the ten tasks we investigate.
\item Further experiments shows that, given strong priors, tree based methods give better results when crucial words are closer to the final representation. If structure priors are unavailable, balanced tree is a good choice, as it makes the path distances between word and sentence encoding to be roughly equal, and in such case, tree encoding can learn the crucial words itself more easily.  
\end{compactitem}

\begin{figure}[t!]
\centering
\subfloat[Encoder-decoder framework for sentence generation.]{
\includegraphics[width=0.2\textwidth]{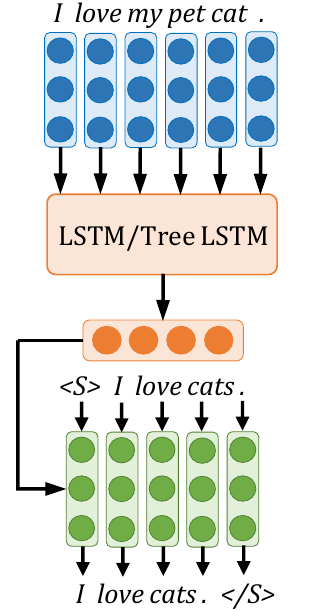}
}
~~~~~
\subfloat[Encoder-classifier framework for sentence classification.] {
\includegraphics[width=0.2\textwidth]{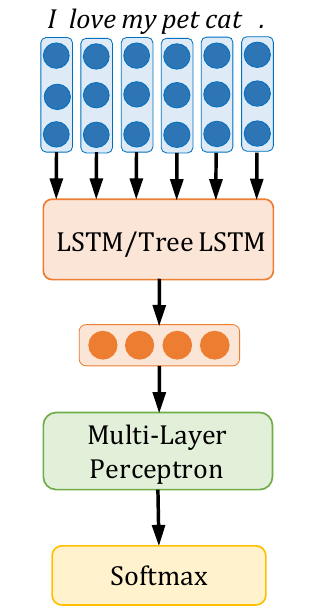}
}
\\
\subfloat[Siamese encoder-classifier framework for sentence relation classification.]{
\includegraphics[width=0.35\textwidth]{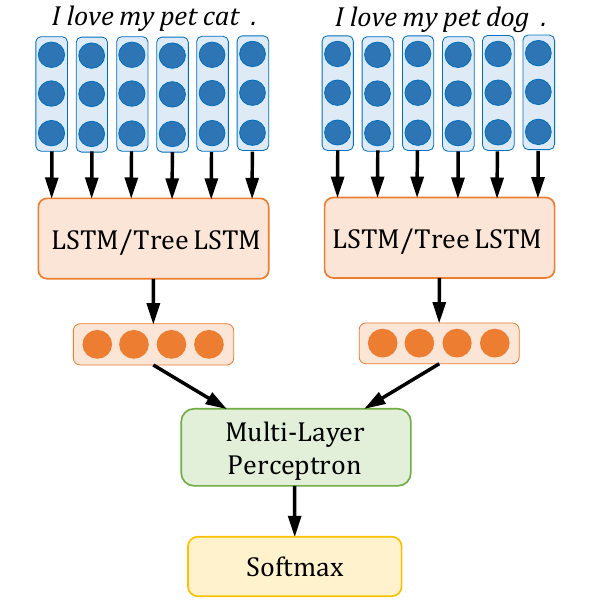}
}
\caption{\label{fig:model-structures} The encoder-classifier/decoder framework for three different groups of tasks. We apply multi-layer perceptron (MLP) for classification, and left-to-right decoders for generation in all experiments. \\[-0.8cm] }
\end{figure}

\section{Experimental Framework}
\label{sec:structures}

\begin{figure*}[t]
    \centering
    \subfloat[Parsing tree.]{
    \includegraphics[width=0.19\textwidth]{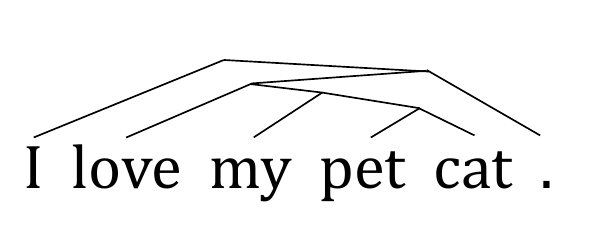}
    }
    \subfloat[Balanced tree.]{
    \includegraphics[width=0.19\textwidth]{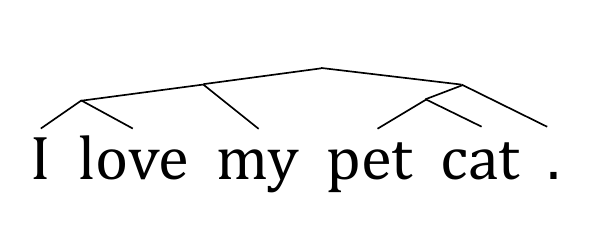}
    }
    \subfloat[Gumbel tree.]{
    \includegraphics[width=0.19\textwidth]{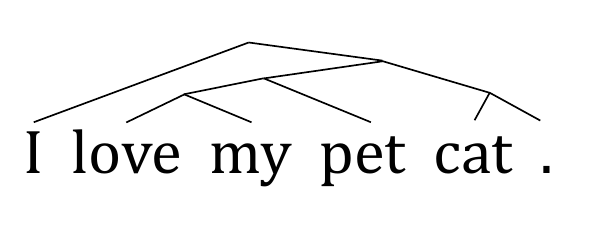}
    }
    \subfloat[Left-branching tree.]{
    \includegraphics[width=0.19\textwidth]{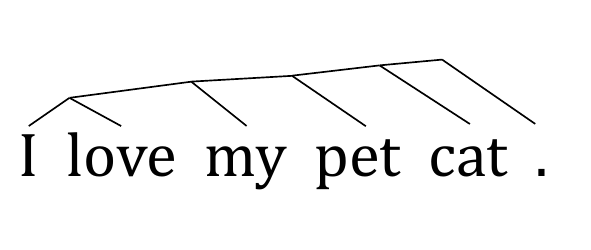}
    }
    \subfloat[Right-branching tree.]{
    \includegraphics[width=0.19\textwidth]{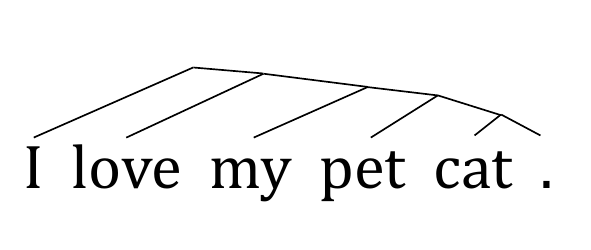}
    }
    \caption{\label{fig:tree-examples} Examples of different tree structures for the encoder part. }
\end{figure*}

We show the applied encoder-classifier/decoder framework for each group of tasks in Figure~\ref{fig:model-structures}.
Our framework has two main components: the encoder part and the classifier/decoder part. 
In general, models encode a sentence to a length-fixed vector, and then applies the vector as the feature for classification and generation. 

We fix the structure of the classifier/decoder, and propose to use five different types of tree structures for the encoder part including:
\begin{compactitem}
    \item Parsing tree. We apply binary constituency tree as the representative, which is widely used in natural language inference \cite{bowman2016fast} and machine translation \cite{eriguchi2016tree,chen2017improved}. Dependency parsing trees~\cite{zhouhao2015,zhouhao2016} are not considered in this paper.
    \item Binary balanced tree. To construct a binary balanced tree, we recursively divide a group of $n$ leafs into two contiguous groups with the size of $\lceil\frac{n}{2}\rceil$ and $\lfloor\frac{n}{2}\rfloor$, until each group has only one leaf node left. 
    \item Gumbel trees, which are produced by straight-forward Gumbel softmax models \cite{choi2017learning}. Note that Gumbel trees are not stable to sentences \cite{williams2018latent}, and we only draw a sample among all of them. 
    \item Left-branching trees. We combine two nodes from left to right, to construct a left-branching tree, which is similar to those generated by the reinforce based RL-SPINN model \cite{williams2018latent}. 
    \item Right-branching trees. In contrast to left-branching ones, nodes are combined from right to left to form a right-branching tree.
\end{compactitem}
We show an intuitive view of the five types of tree structures in Figure~\ref{fig:tree-examples}.
In addition, existing works \cite{choi2017learning,williams2018latent} show that using hidden states of bidirectional RNNs as leaf node representations (bi-leaf-RNN) instead of word embeddings may improve the performance of tree LSTMs, as leaf RNNs help encode context information more completely. 
Our framework also support leaf RNNs for tree LSTMs.

\section{Description of Investigated Tasks}
\label{sec:tasks}
We conduct experiments on 10 different tasks, which are grouped into 3 categories, namely the single sentence classification (5 tasks), sentence relation classification (2 tasks), and sentence generation (3 tasks).
Each of the tasks is compatible to the encoder-classifier/decoder framework shown in Figure~\ref{fig:model-structures}. 
These tasks cover a wide range of NLP applications, and depend on different granularities of features.

Note that the datasets may use articles or paragraphs as instances, some of which consist of only one sentence. 
For each dataset, we only pick the subset of single-sentence instances for our experiments, and the detailed meta-data is in Table~\ref{table:metadata}. 

\subsection{Sentence Classification}
First, we introduce four text classification datasets from \newcite{zhang2015character}, including AG's News, Amazon Review Polarity , Amazon Review Full and DBpedia. 
Additionally, noticing that parsing tree was shown to be effective \cite{li2015tree} on the task of word-level semantic relation classification \cite{hendrickx2009semeval}, we also add this dataset to our selections.  

\paragraph{AG's News (AGN).} 
Each sample in this dataset is an article, associated with a label indicating its topic: world, sports, business or sci/tech. 

\paragraph{Amazon Review Polarity (ARP).}
The Amazon Review dataset is obtained from the Stanford Network Analysis Project (SNAP; \citeauthor{mcauley2013hidden}, \citeyear{mcauley2013hidden}). 
It collects a large amount of product reviews as paragraphs, associated with a star rate from 1 (most negative) to 5 (most positive).
In this dataset, 3-star reviews are dropped, while others are classified into two groups: positive (4 or 5 stars) and negative (1 or 2 stars). 

\paragraph{Amazon Review Full (ARF).}
Similar to the ARP dataset, the ARF dataset is also collected from Amazon product reviews. 
Labels in this dataset are integers from 1 to 5. 

\paragraph{DBpedia.}
DBpedia is a crowd-sourced community effort to extract structured
information from Wikipedia \cite{lehmann2015dbpedia}. 
\newcite{zhang2015character} select 14 non-overlapping
classes from DBpedia 2014 to construct this dataset. 
Each sample is given by the title and abstract of the Wikipedia article, associated with the class label.

\paragraph{Word-Level Semantic Relation (WSR)}
SemEval-2010 Task 8 \cite{hendrickx2009semeval} is to find semantic relationships between pairs of nominals.
Each sample is given by a sentence, of which two nominals are explicitly indicated, associated with manually labeled semantic relation between the two nominals.
For example, the sentence ``\textit{My [apartment]$_{e_1}$ has a pretty large [kitchen]$_{e_2}$ .}'' has the label \textit{component-whole}($e_2$, $e_1$).
Different from retrieving the path between two labels \cite{li2015tree,socher2013recursive}, we feed the entire sentence together with the nominal indicators (\textit{i.e.}, tags of $e_1$ and $e_2$) as words to the framework.
We also ignore the order of $e_1$ and $e_2$ in the labels given by the dataset.
Thus, this task turns to be a 10-way classification one. 

\subsection{Sentence Relation Classification}
To evaluate how well a model can capture semantic relation between sentences, we introduce the second group of tasks: sentence relation classification.
\paragraph{Natural Language Inference (NLI).}
The Stanford Natural Language Inference (SNLI) Corpus \cite{bowman2015large} is a challenging dataset for sentence-level textual entailment. 
It has 550K training sentence pairs, as well as 10K for development and 10K for test. 
Each pair consists of two relative sentences, associated with a label which is one of \textit{entailment, contradiction} and \textit{neutral}. 
\paragraph{Conjunction Prediction (Conj).}
Information about the coherence relation between two sentences is sometimes apparent in the text explicitly \cite{miltsakaki2004penn}: this is the case whenever the second sentence starts with a conjunction phrase.
\newcite{jernite2017discourse} propose a method to create \textit{conjunction prediction} dataset from unlabeled corpus. 
They create a list of phrases, which can be classified into nine types, as conjunction indicators.
The object of this task is to recover the conjunction type of given two sentences, which can be used to evaluate how well a model captures the semantic meaning of sentences.
We apply the method proposed by~\newcite{jernite2017discourse} on the Wikipedia corpus to create our conj dataset.

\subsection{Sentence Generation}
We also include the sentence generation tasks in our experiments, to investigate the representation ability of different encoders over global (long-term) context features. 
Note that our framework is based on encoding, which is different from those attention based approaches. 
\paragraph{Paraphrasing (Para).}
Quora Question Pair Dataset is a widely applied dataset to evaluate  paraphrasing models \cite{wang2017bilateral,li2017paraphrase}. \footnote{\url{https://data.quora.com/First-Quora-Dataset-Release-Question-Pairs}}
In this work, we treat the paraphrasing task as a sequence-to-sequence one, and evaluate on it with our sentence generation framework.

\paragraph{Machine Translation (MT).} 
Machine translation, especially cross-language-family machine translation, is a complex task, which requires models to capture the semantic meanings of sentences well.
We apply a large challenging English-Chinese sentence translation task for this investigation, which is adopted by a variety of neural translation work \cite{tu2016modeling,li2017modeling,chen2017improved}.
We extract the parallel data from the LDC corpora,\footnote{The corpora includes LDC2002E18, LDC2003E07, LDC2003E14, Hansards portion of LDC2004T07, LDC2004T08 and LDC2005T06} selecting 1.2M from them as our training set, 20K and 80K of them as our development set and test set, respectively.

\paragraph{Auto-Encoding (AE).}
We extract the English part of the machine translation dataset to form a auto-encoding task, which is also compatible with our encoder-decoder framework.

\begin{table}[t]
\centering
\begin{tabular}{lrrrrr}
\toprule 
\textbf{Dataset} & \multicolumn{3}{c}{\textbf{\#Sentence}} & \textbf{\#Cls} & \textbf{Avg.} \\
&       \small Train & \small Dev & \small Test &                        &  \textbf{Len}\\
\midrule
\multicolumn{6}{l}{\textit{Sentence Classification}} \\
\midrule
News		& 60K  & 6.7K & 4.3K	& 4  & 31.5 \\
ARP  	    & 128K & 14K  & 16K	    & 2  & 33.7 \\
ARF	        & 110K & 12K  & 27K 	& 5  & 33.8 \\
DBpedia		& 106K & 11K  & 15K 	& 14 & 20.1 \\
WSR		& 7.1K & 891  & 2.7K	& 10 & 23.1 \\
\midrule
\multicolumn{6}{l}{\textit{Sentence Relation }} \\
\midrule
SNLI 		& 550K & 10K & 10K 	& 3 & 11.2 \\
Conj		& 552K & 10K & 10K	& 9 & 23.3 \\
\midrule
\multicolumn{6}{l}{\textit{Sentence Generation}} \\
\midrule
Para	    & 98K & 2K & 3K	    & N/A & 10.2 \\
MT			& 1.2M & 20K & 80K	& N/A & 34.1 \\
AE			& 1.2M & 20K & 80K	& N/A & 34.1 \\
\bottomrule
\end{tabular}
\caption{\label{table:metadata} Meta-data of the downstream tasks we investigated. 
For each task, we list the quantity of instances in train/dev/test set, the average length (by words) of sentences (source sentence only for generation task), as well as the number of classes if applicable.}
\end{table}

\section{Experiments}
\label{sec:exprs}
\begin{table*}[t]
\centering
\begin{tabular}{l|rrrrr|rr|rrr}
     \toprule
     & \multicolumn{5}{c|}{\textit{\small Sentence Classification}} &\multicolumn{2}{c}{\textit{\small Sentence Relation}} &\multicolumn{3}{|c}{\small\textit{Sentence Generation}}\\
     \textbf{Model} & \textbf{AGN} & \textbf{ARP} & \textbf{ARF} & \textbf{DBpedia} & \textbf{WSR} & \textbf{NLI} & \textbf{Conj} & \textbf{Para} & \textbf{MT} & \textbf{AE} \\
     \midrule
     \multicolumn{11}{l}{\textit{{Latent Trees}}} \\
     \midrule
     Gumbel & 91.8  & 87.1  & 48.4  & 98.6  & 66.7  &  80.4  & 51.2  & 20.4  &  17.4  &  39.5 \\
     \multicolumn{1}{r|}{\textit{+bi-leaf-RNN}}  & 91.8  & \textbf{88.1}  & \textbf{49.7}  & 98.7  & 69.2  & \textbf{82.9}  & 53.7  & 20.5  & 22.3  & 75.3  \\
     \midrule
     \multicolumn{11}{l}{\textit{{(Constituency) Parsing Trees}}} \\
     \midrule
     Parsing & 91.9  & 87.5  & 49.4  & \textbf{98.8}  & 66.6  & 81.3  & 52.4  & 19.9  & 19.1  & 44.3  \\
     \multicolumn{1}{r|}{\textit{+bi-leaf-RNN}}  & 92.0  & 88.0  & 49.6  & \textbf{98.8}  & 68.6  & 82.8  & 53.4  & 20.4  & 22.2  & 72.9  \\
     \midrule
     \multicolumn{11}{l}{\textit{{Trivial Trees}}} \\
     \midrule
     Balanced & 92.0  & 87.7  & 49.1  & 98.7  & 66.2  & 81.1  &  52.1 & 19.7  & 19.0  & 49.4  \\
     \multicolumn{1}{r|}{\textit{+bi-leaf-RNN}}  & \textbf{92.1}  & 87.8  & \textbf{49.7}  & \textbf{98.8}  & \textbf{69.6}  & 82.6  & \textbf{54.0}  & 20.5  & 22.3  & 76.0  \\
     Left-branching & 91.9  & 87.6  & 48.5  & 98.7  & 67.8  & 81.3  & 50.9  & 19.9  & 19.2  &  48.0 \\
     \multicolumn{1}{r|}{\textit{+bi-leaf-RNN}}  & 91.2  & 87.6  & 48.9  & 98.6  & 67.7  & 82.8  & 53.3  & 20.6  & 21.6  & 72.9  \\
     Right-branching & 91.9  & 87.7  & 49.0  & \textbf{98.8}  & 68.6 & 81.0  & 51.3  &  20.4  &  19.7  &  54.7\\
     \multicolumn{1}{r|}{\textit{+bi-leaf-RNN}}  & 91.9 & 87.9  & 49.4 & 98.7  & 68.7  & 82.8 & 53.5  & \bf 20.9 & \bf 23.1  & \bf 80.4  \\
     \midrule
     \multicolumn{11}{l}{\textit{{Linear Structures}}} \\
     \midrule
     LSTM                    & 91.7  & 87.8  & 48.8  & 98.6  & 66.1  & 82.6  & 52.8  & 20.3  & 19.1  & 46.9  \\
     \multicolumn{1}{r|}{\textit{+bidirectional}} & 91.7  & 87.8  & 49.2  & 98.7  & 67.4  & 82.8  & 53.3  & 20.2  & 21.3  & 67.0  \\
     \midrule
     \textbf{Avg. Length}    & \underline{31.5} & \underline{33.7} & \underline{33.8} & \underline{20.1} & \underline{23.1} & \underline{11.2} & \underline{23.3} & \underline{10.2} & \underline{34.1} & \underline{34.1} \\
     \bottomrule
\end{tabular}
\caption{\label{table:main-result} 
Test results for different encoder architectures trained by a unified encoder-classifier/decoder framework. 
We report accuracy $(\times 100)$ for classification tasks, and BLEU score (\citeauthor{papineni2002bleu}, \citeyear{papineni2002bleu}; word-level for English targets and char-level for Chinese targets) for generation tasks. 
Large is better for both of the metrics.
The best number(s) for each task are in bold.
In addition, average sentence length (in words) of each dataset is attached in the last row with underline.  \\[-0.8cm]}
\end{table*}

In this section, we present our experimental results and analysis. 
Section~\ref{sec:set-up} introduces our set-up for all the experiments.
Section~\ref{sec:main-results} shows the main results and analysis on ten downstream tasks grouped into three classes, which can cover a wide range of NLP applications. 
Regarding that trivial tree based LSTMs perform the best among all models, we draw two hypotheses, which are i) right-branching tree benefits a lot from strong structural priors; ii) balanced tree wins because it fairly treats all words  so that crucial information could be more easily learned by the LSTM gates automatically. We test the hypotheses in Section~\ref{sec:trivial}. 
Finally, we compare the performance of linear and tree LSTMs with three widely applied pooling mechanisms in Section~\ref{sec:pooling}.

\subsection{Set-up}
\label{sec:set-up}
In experiments, we fix the structure of the classifier as a two-layer MLP with ReLU activation, and the structure of decoder as GRU-based recurrent neural networks \cite{cho2014learning}. \footnote{We observe that ReLU can significantly boost the performance of Bi-LSTM on SNLI.} 
The hidden-layer size of MLP is fixed to 1024, while that of GRU is adapted from the size of sentence encoding. 
We initialize the word embeddings with 300-dimensional {\tt GloVe} \cite{pennington2014glove} vectors.\footnote{\url{http://nlp.stanford.edu/data/glove.840B.300d.zip}}
We apply 300-dimensional bidirectional (600-dimensional in total) LSTM as leaf RNN when necessary.
We use Adam \cite{kingma2014adam} optimizer to train all the models, with the learning rate of 1e-3 and batch size of 64. 
In the training stage, we drop the samples with the length of either source sentence or target sentence larger than 64. 
We do not apply any regularization or dropout term in all experiments except the task of WSR, on which we tune dropout term with respect to the development set. 
We generate the binary parsing tree for the datasets without parsing trees using ZPar \cite{zhang2011syntactic}.\footnote{\url{https://www.sutd.edu.sg/cmsresource/faculty/yuezhang/zpar.html}}
More details are summarized in supplementary materials. 

\subsection{Main Results}
\label{sec:main-results}
In this subsection, we aim to compare the results from different encoders. We do not include any attention \cite{wang2016attention,lin2017astructured} or pooling \cite{collobert2008unified,socher2011dynamic,zhou2016text} mechanism here, in order to avoid distractions and make the encoder structure affects the most. 
We will further analyze pooling mechanisms in Section \ref{sec:pooling}.

Table~\ref{table:main-result} presents the performances of different encoders on a variety of downstream tasks, which lead to the following observations:
\paragraph{Tree encoders are useful on some tasks.} We get the same conclusion with \newcite{li2015tree} that tree-based encoders perform better on tasks requiring long-term context features.
Despiting the linear structured left-branching and right-branching tree encoders, we find that, tree-based encoders generally perform better than Bi-LSTMs on tasks of \textit{sentence relation} and \textit{sentence generation}, which may require relatively more long term context features for obtaining better performances.
However, the improvements of tree encoders on \textbf{NLI} and \textbf{Para} are relatively small, which may be caused by that sentences of the two tasks are shorter than others, and the tree encoder does not get enough advantages to capture long-term context in short sentences.

\paragraph{Trivial tree encoders outperform other encoders.}
Surprisingly, binary balanced tree encoder gets the best results on most tasks of classification and right-branching tree encoder tends to be the best on sentence generation.
Note that binary balanced tree and right-branching tree are only trivial tree structures, but outperform syntactic tree and latent tree encoders.
The latent tree is really competitive on some tasks, as its  structure is directly tuned by the corresponding tasks.
However, it only beats the binary balanced tree by very small margins on NLI and ARP.
We will give analysis about this in Section \ref{sec:trivial}.

\paragraph{Larger quantity of parameters is not the only reason of the improvements.}
Table \ref{table:main-result} shows that tree encoders benefit a lot from adding leaf-LSTM, which brings not only sentence level information to leaf nodes, but also more parameters than the bi-LSTM encoder.
However, left-branching tree LSTM has a quite similar structure with linear LSTM, and it can be viewed as a linear LSTM-on-LSTM structure. 
It has the same amounts of parameters as other tree-based encoders, but still falls behind the balance tree encoder on most of the tasks.
This indicates that larger quantity of parameters is at least not the only reason for binary balance tree LSTM encoders to gain improvements against bi-LSTMs.

\subsection{Why Trivial Trees Work Better?}
\label{sec:trivial}

Binary balanced tree and right-branching are trivial ones, hardly containing syntax information.
In this section, we analyze why these trees achieve high scores in deep.

\subsubsection{Right Branching Tree Benefits from Strong Structural Prior}
\label{sec:right-better}
\begin{figure*}[t]
    \centering
    \subfloat[Balanced tree, MT.]{
    \includegraphics[width=0.24\textwidth]{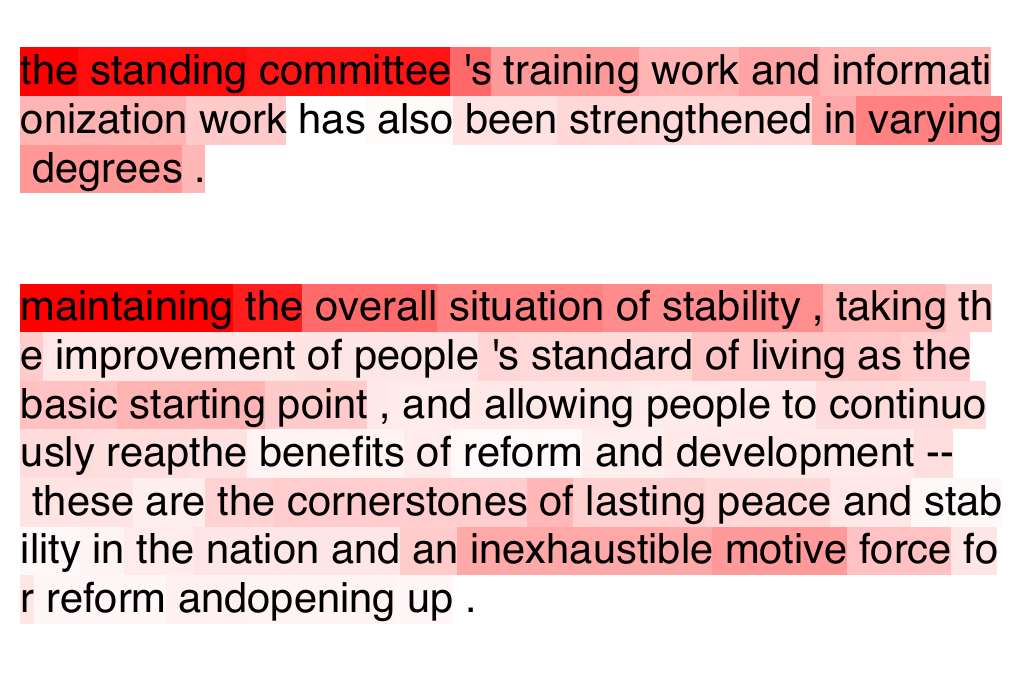}
    }
    \subfloat[Left-branching tree, MT.]{
    \includegraphics[width=0.24\textwidth]{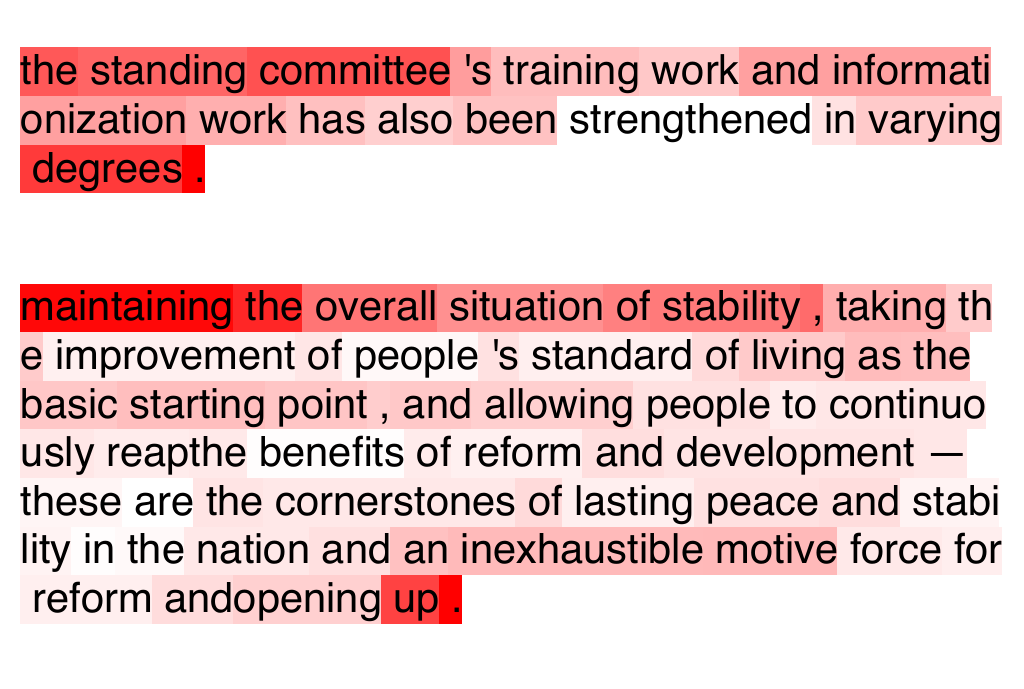}
    }
      \subfloat[Right-branching, MT.]{
    \includegraphics[width=0.24\textwidth]{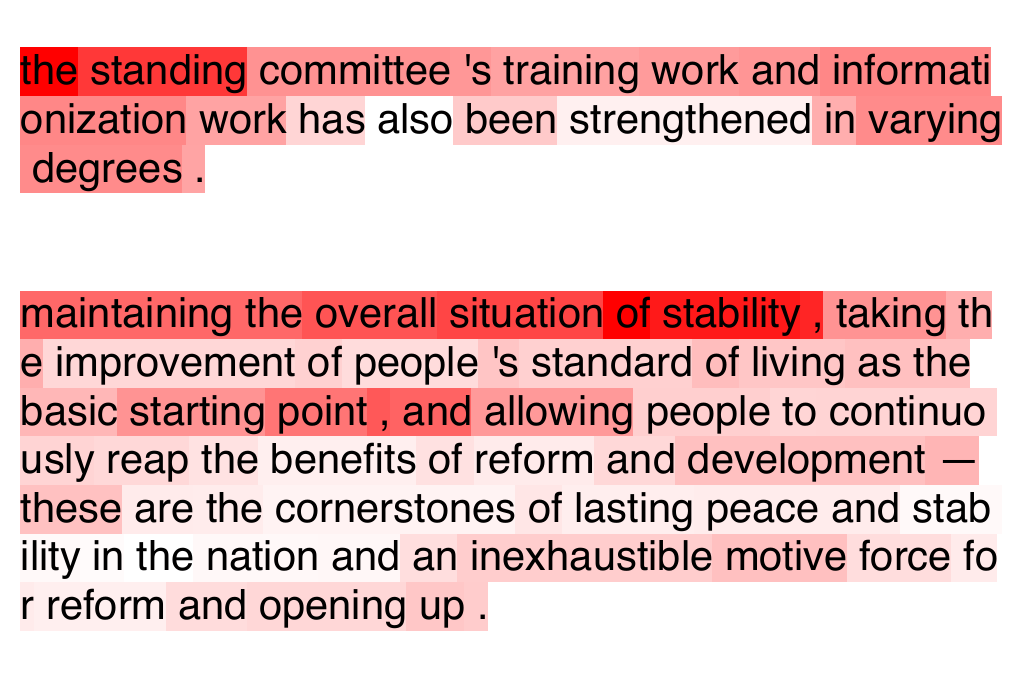}
    } 
    \subfloat[Bi-LSTM, MT.]{
    \includegraphics[width=0.24\textwidth]{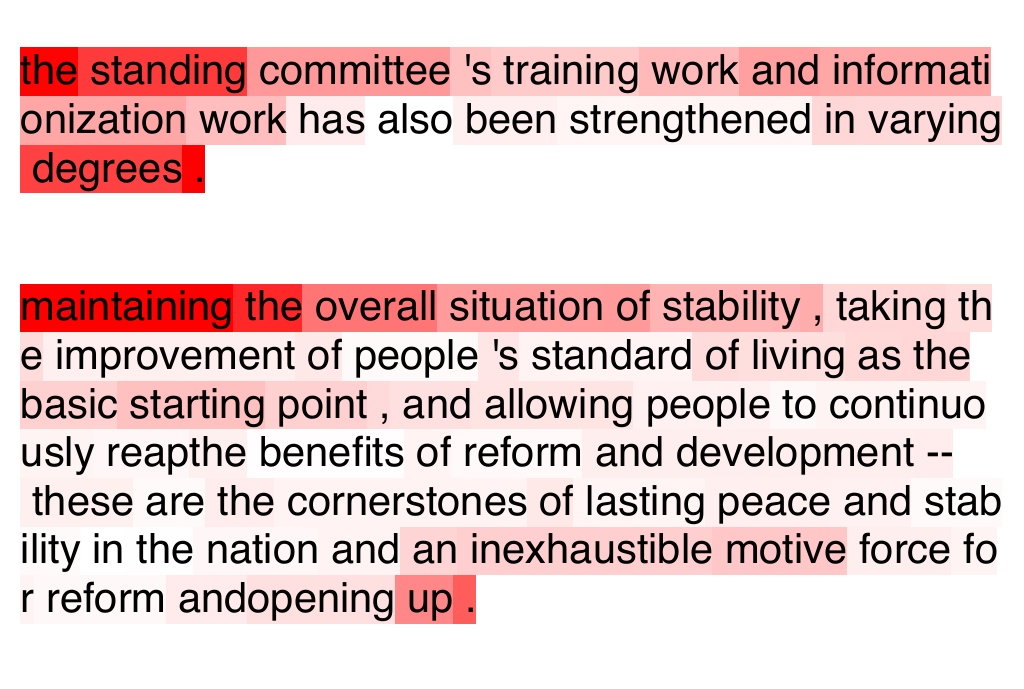}
    }
    
    \subfloat[Balanced tree, AE.]{
    \includegraphics[width=0.24\textwidth]{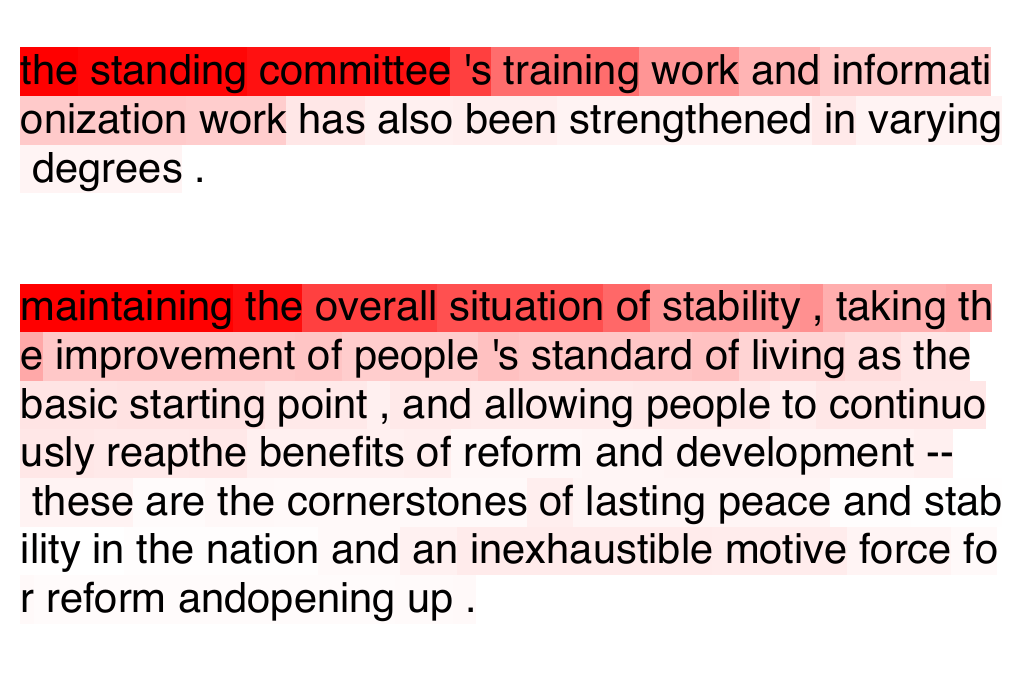}
    }
        \subfloat[Left-branching tree, AE.]{
    \includegraphics[width=0.24\textwidth]{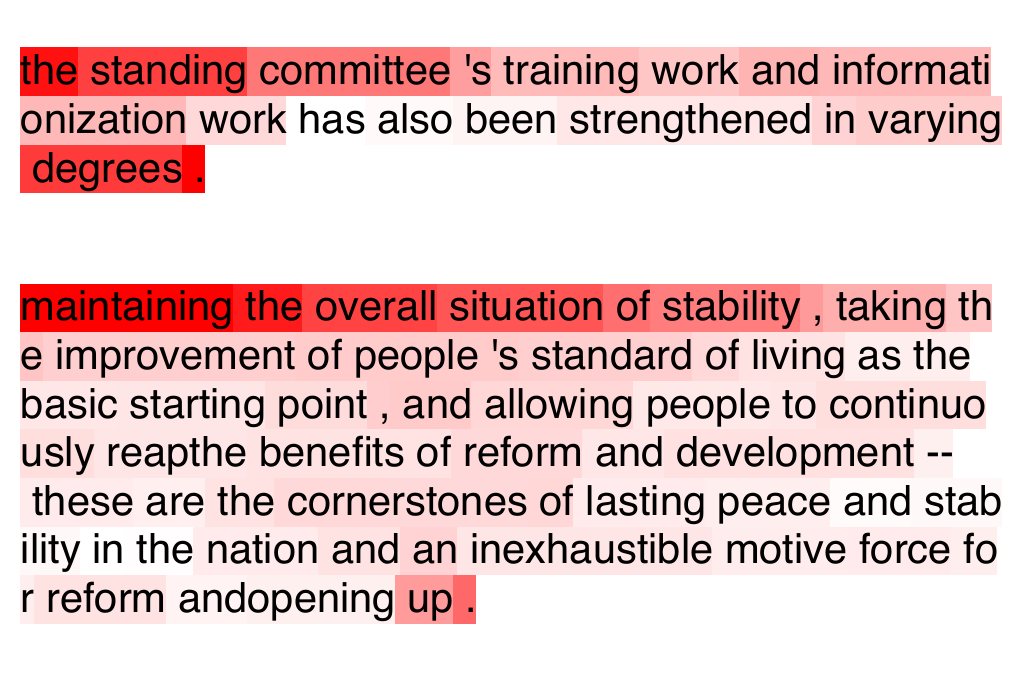}
    }
          \subfloat[Right-branching, AE.]{
    \includegraphics[width=0.24\textwidth]{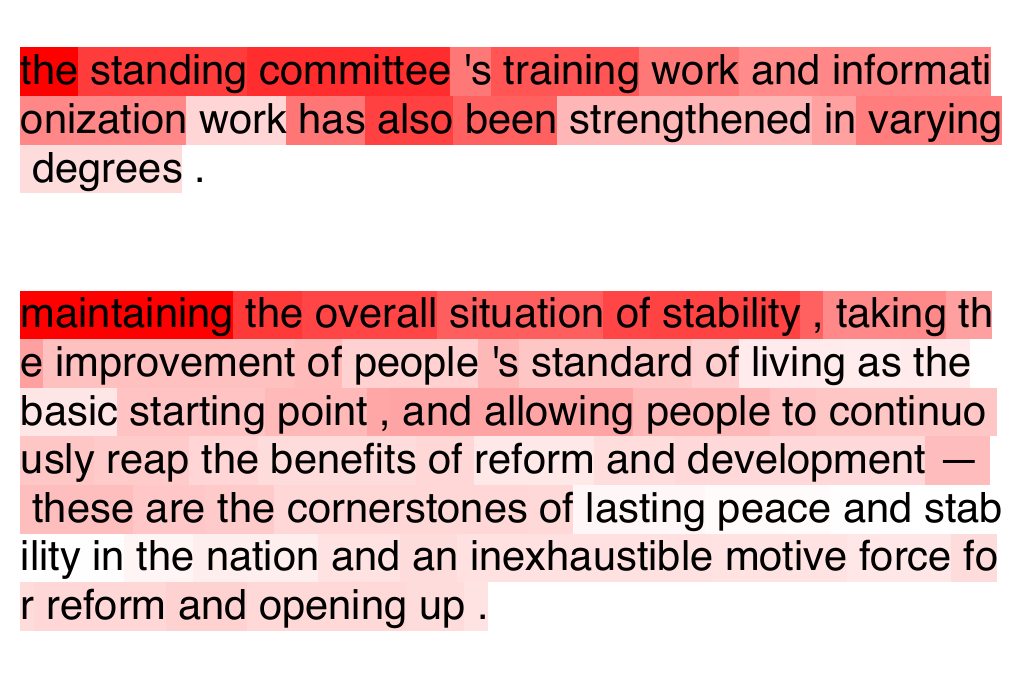}
    }
    \subfloat[Bi-LSTM, AE.]{
     \includegraphics[width=0.24\textwidth]{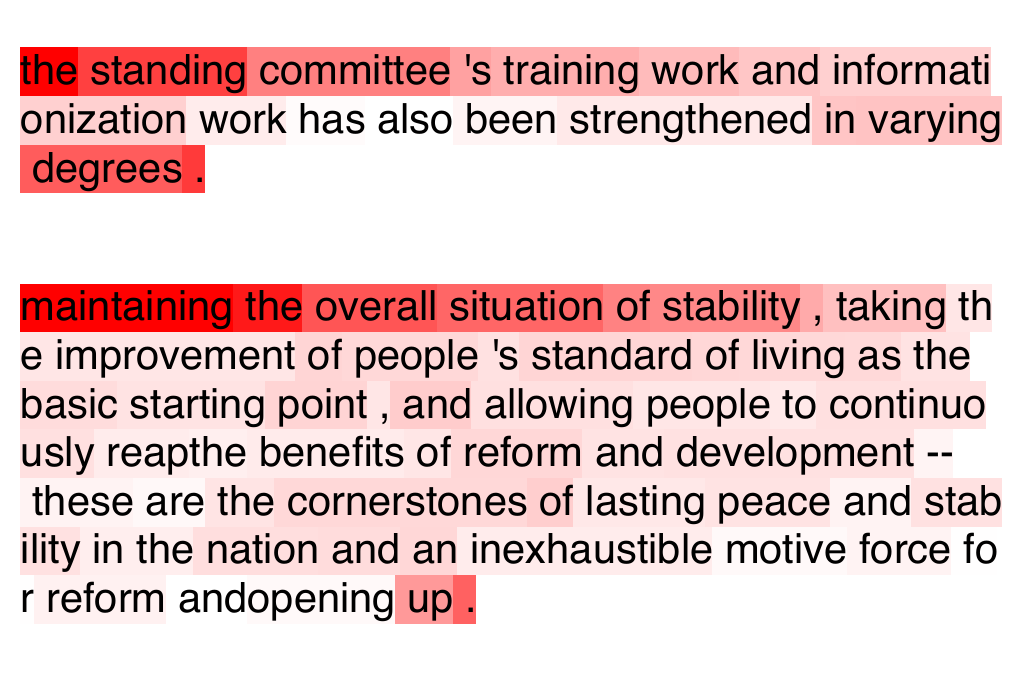}
    }
    \caption{\label{fig:saliency} Saliency visualization of words in learned MT and AE models. Darker means more important to the sentence encoding. }
\end{figure*}

We argue that right-branching trees benefit from its strong structural prior.
In sentence generation tasks, models generate sentences from left to right, which makes words in the left of the source sentence more important~\cite{sutskever2014sequence}.
If the encoder fails to memorize the left words, the information about right words would not help due to the error propagation.
In right-branching trees, left words of the sentence are closer to the final representation, which makes the left words are more easy to  be memorized, and we call this structure prior.
Oppositely, in the case of  left-branching trees, right words of the sentence are closer to the representation.

To validate our hypothesis, we propose to visualize the Jacobian as word-level saliency \cite{shi2018learning}, which can be viewed as the contribution of each word to the sentence encoding: 
\begin{equation*}
    \mathrm{J}(\mathbf{s}, \mathbf{w}) = \lVert\nabla \mathbf{s}(\mathbf{w})\rVert_1 = \sum_{i,j} \vert\frac{\partial s_i}{\partial w_j}\vert 
\end{equation*}
where $\mathbf{s}=(s_1, s_2, \cdots, s_p)^T$ denotes the embedding of a sentence, and $\mathbf{w}=(w_1, w_2, \cdots, w_q)^T$ denotes embedding of a word.
We can compute the saliency score using backward propagation.
For a word in a sentence, higher saliency score means more contribution to  sentence encoding. 

We present the visualization in Figure~\ref{fig:saliency} using the visualization tool from \newcite{lin2017astructured}.
It shows that right-branching tree LSTM encoders tend to look at the left part of the sentence, which is very helpful to the final generation performance, as left words are more crucial.
Balanced trees also have this feature and we think it is because balance tree treats these words fairly, and crucial information could be more easily learned by the LSTM gates automatically.

However, bi-LSTM and left-branching tree LSTM also pay much attention to words in the right~(especially the last two words), which maybe caused by the short path from the right words to the root representation, in the two corresponding tree structures. 

Additionally, Table~\ref{table:pearson} shows that models trained with the same hyper-parameters but different initializations have strong agreement with each other. 
Thus, ``looking at the first words'' is a stable behavior of balanced and right-branching tree LSTM encoders in sentence generation tasks. So is ``looking at the first and the last words'' for Bi-LSTMs and left-branching tree LSTMs.

\begin{table}[t]
    \centering
    \begin{tabular}{lrr}
         \toprule
         \textbf{Model} & \textbf{MT} & \textbf{AE} \\
         \midrule
         Balanced (BiLRNN)  & 93.1  & 96.9 \\
         Left-Branching (BiLRNN) & 94.2 & 95.4 \\
         Right-Branching (BiLRNN) & 92.3 & 95.1 \\
         Bi-LSTM            & 96.4  & 96.1 \\
         \bottomrule
    \end{tabular}
    \caption{\label{table:pearson} Mean average Pearson correlation $(\times 100)$ across five models trained with same hyper-parameters. For each testing sentence, we compute the saliency scores of words. Cross-model Pearson correlation can show the agreement of two models on one sentence, and average Pearson correlation is computed through all sentences. We report mean average Pearson correlation of the $5 \times 4$ model pairs. }
\end{table}

\subsubsection{Binary Balanced Tree Benefits from Shallowness}
\label{sec:balanced-better}
\begin{figure}[t!]
    \centering
    \subfloat[$\rho$-depth line for WSR.]{
    \includegraphics[width=0.21\textwidth]{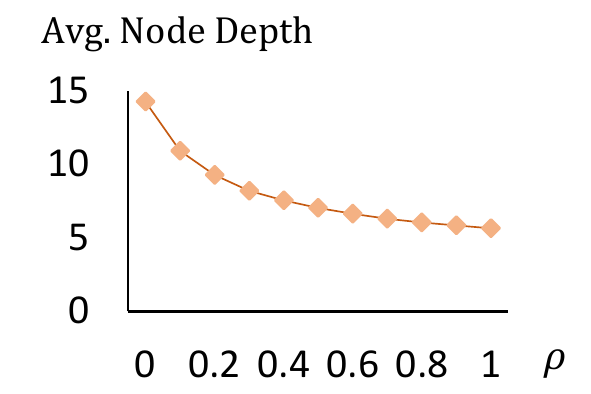}
    }
    ~~\subfloat[$\rho$-Acc. line for WSR.]{
    \includegraphics[width=0.21\textwidth]{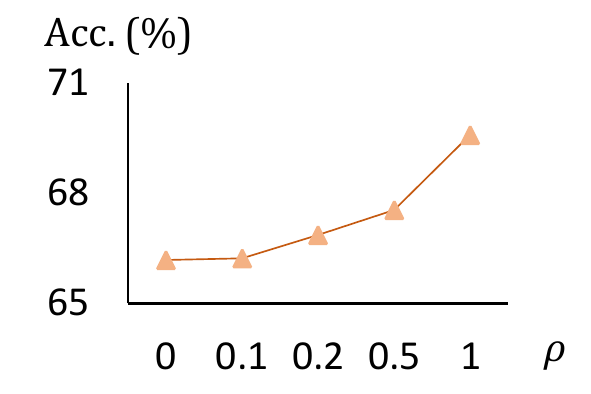}
    }
    \\
    \subfloat[$\rho$-depth line for MT.]{
    \includegraphics[width=0.21\textwidth]{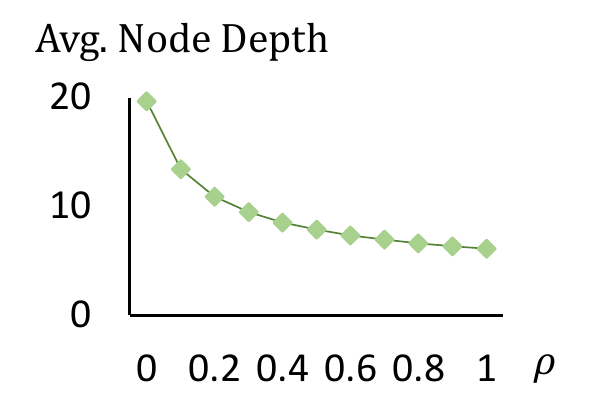}
    }
    ~~\subfloat[$\rho$-BLEU line for MT.]{
    \includegraphics[width=0.21\textwidth]{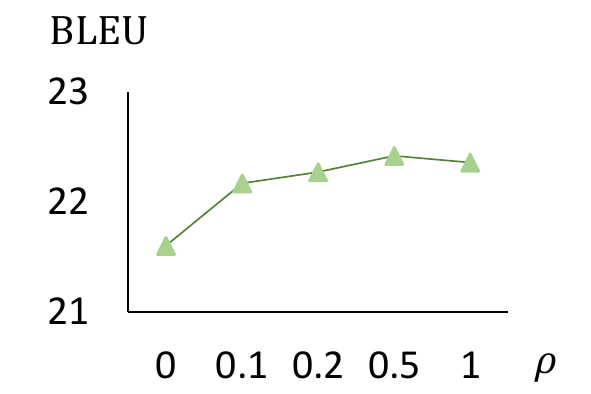}
    }
    \\
    \subfloat[$\rho$-depth line for AE.]{
    \includegraphics[width=0.21\textwidth]{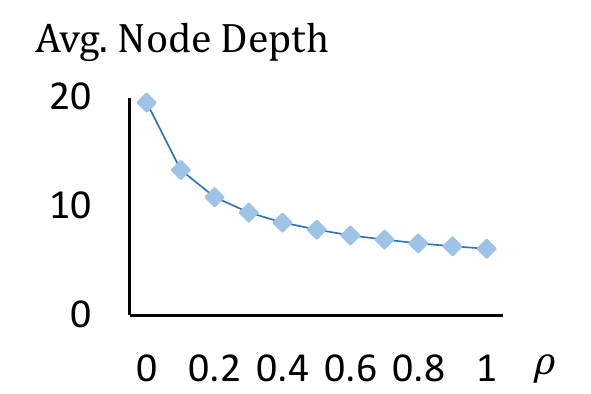}
    }
    ~~\subfloat[$\rho$-BLEU line for AE.]{
    \includegraphics[width=0.21\textwidth]{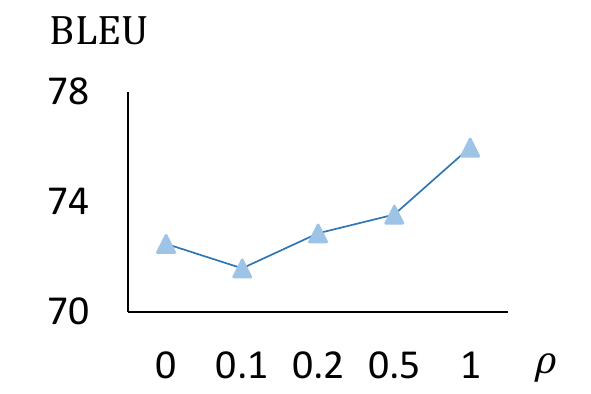}
    }
    \caption{\label{fig:tree-depth-performance} $\rho$-depth and $\rho$-performance lines for three tasks. 
    There is a trend that the depth drops and the performance raises with the growth of $\rho$. }
\end{figure}   

\begin{figure*}[t!]
    \centering
    \subfloat[Length-Accuracy lines for WSR.]{
    \includegraphics[width=0.3\textwidth]{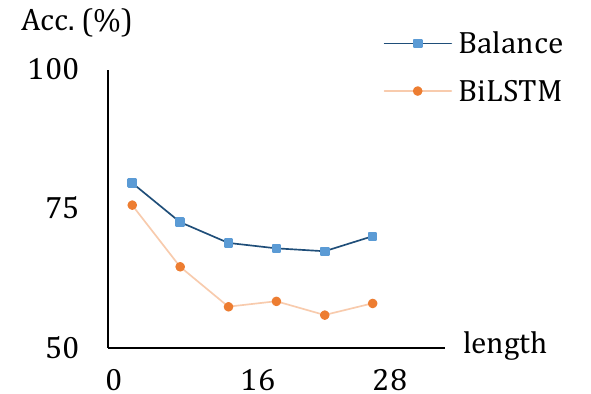}
    }
    \subfloat[Length-BLEU lines for MT.]{
    	\includegraphics[width=0.3\textwidth]{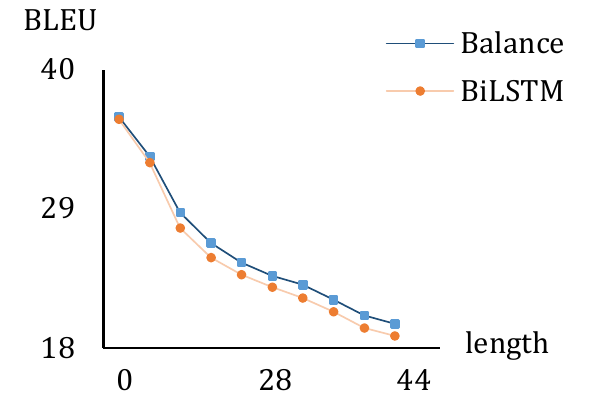}
    }
    \subfloat[Length-BLEU lines for AE.]{
        \includegraphics[width=0.3\textwidth]{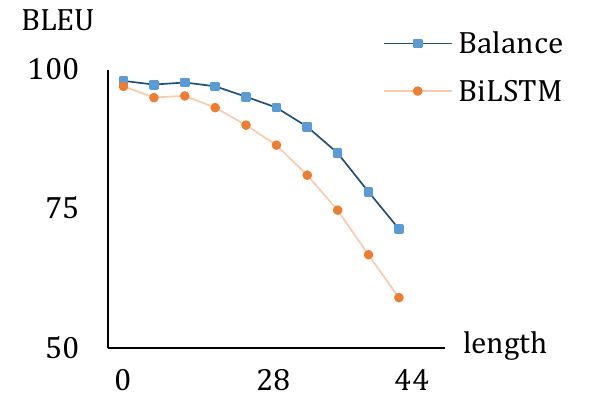}
    }
    \caption{\label{fig:length-performance} Length-performance lines for the further investigated tasks. We divide test instances into several groups by length, and report the performance on each group respectively. Sentences with length in $[1,8]$ are put to the first group, and the group $i (i\geq 2)$ covers the range of $[4i+1, 4i+4]$ in length. ]}
\end{figure*}

Compared to syntactic and latent trees, the only advantage of balanced tree we can hypothesize is that, it is shallower and more balanced than others.
Shallowness may lead to shorter path for information propagation from leafs to the root representation, and makes the representation learning more easy due to the reduction of errors in the propagation process.
Balance makes the tree fairly treats all leaf nodes, which makes it more easily to automatically select the crucial information over all words in a sentence.  

To test our hypothesis, we conduct the following experiments. 
We select three tasks, on which binary balanced tree encoder wins Bi-LSTMs with a large margin~(WSR, MT and AE).
We generate random binary trees for sentences, while controlling the depth using a hyper-parameter $\rho$. 
We start by a group with all words (nodes) in the sentence.
At each time, we separate $n$ nodes to two continuous groups sized $(\lceil \frac{n}{2} \rceil$, $\lfloor \frac{n}{2} \rfloor)$ with probability $\rho$, while those sized $(n-1, 1)$ with probability $1-\rho$.
Trees generated with $\rho=0$ are exactly left-branching trees, and those generated with $\rho=1$ are binary balanced trees. 
The expected node depth of the tree turns smaller with $\rho$ varies from 0 to 1.

Figure~\ref{fig:tree-depth-performance} shows that, in general, trees with shallower node depth have better performance on all of the three tasks (for binary tree, shallower also means more balanced), which validates our above hypothesis that binary balanced tree gains the reward from its shallow and balanced structures. 

Additionally, Figure~\ref{fig:length-performance} demonstrates that binary balanced trees work especially better with relative long sentences. 
As desired, on short-sentence groups, the performance gap between Bi-LSTM and binary balanced tree LSTM is not obvious, while it grows with the test sentences turning longer. 
This explains why tree-based encoder gives small improvements on NLI and Para, because sentences on these two tasks are much shorter than others.

\subsection{Can Pooling Replace Tree Encoder?}
\label{sec:pooling}
\begin{figure}[t]
\centering
\subfloat[Balanced tree.]{
\includegraphics[width=0.4\textwidth]{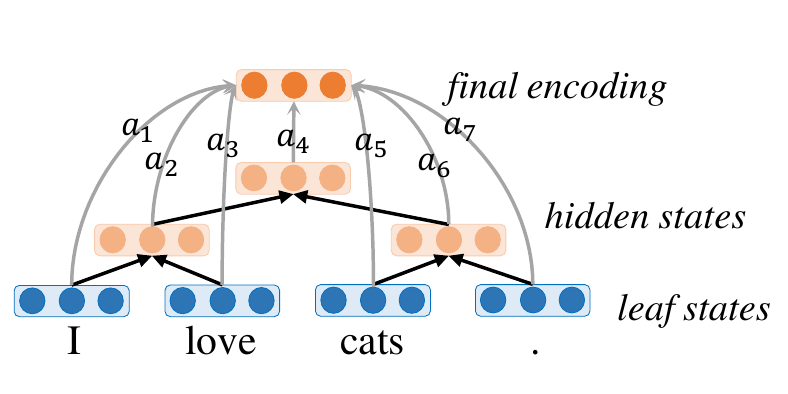}
}

\subfloat[Bi-LSTM.]{
\includegraphics[width=0.4\textwidth]{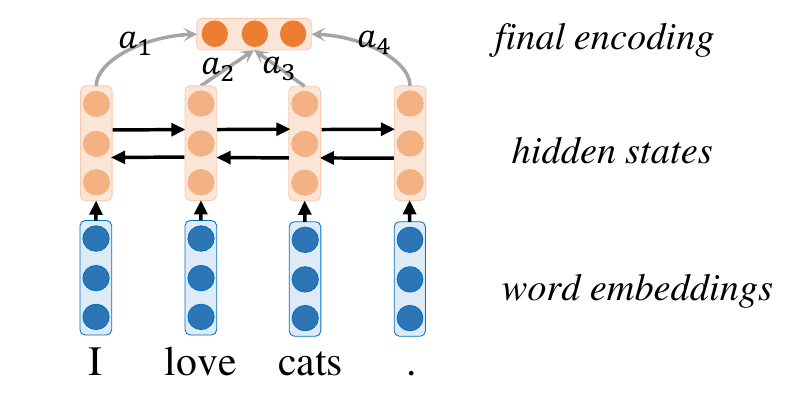}
}
\caption{\label{fig:attention} An illustration of the investigated self-attentive pooling mechanism. \\[-0.8cm]}
\end{figure}

Max pooling \cite{collobert2008unified,zhao2015self}, mean pooling \cite{conneau2017supervised} and self-attentive pooling \cite[also known as self-attention;][]{santos2016attentive,liu2016learning,lin2017astructured} are three popular and efficient choices to improve sentence encoding.
In this part, we will compare the performance of tree LSTMs and bi-LSTM on the tasks of WSR, MT and AE, with each pooling mechanism respectively, aiming to demonstrate the role that pooling plays in sentence modeling, and validate whether tree encoders can be replaced by pooling. 

As shown in Figure~\ref{fig:attention}, for linear LSTMs, we apply pooling mechanism to all hidden states; as for tree LSTMs, pooling is applied to all hidden states and leaf states of tree LSTMs. Implementation details are summarized in the supplementary materials.

Table \ref{table:pooling} shows that max and attentive pooling improve all the structures on the task of WSR, but all the pooling mechanisms fail on MT and AE that require the encoding to capture complete information of sentences, while pooling mechanism may cause the loss of information through the procedure. 
The result indicates that, though pooling mechanism is efficient on some tasks, it cannot totally gain the advantages brought by tree structures. Additionally, we think the attention mechanism has the benefits of the balanced tree modeling, which also fairly treat all words and learn the crucial parts automatically. The path from representation to words in attention are even shorter than the balanced tree. Thus the fact that attentive pooling outperforms balanced trees on WSR is not surprising to us.

\begin{table}[t]
    \centering
    \begin{tabular}{lrrr}
        \toprule
         \textbf{Model} & \textbf{WSR} & \textbf{MT} & \textbf{AE} \\
         \midrule
         Bi-LSTM                        & 67.4  & 21.3  & 67.0  \\
         \multicolumn{1}{r}{\textit{+max-pooling}}      & 71.8 $\uparrow$ & 21.6 $\uparrow$ & 48.0  $\downarrow$ \\
         \multicolumn{1}{r}{\textit{+mean-pooling}}      & 64.3 $\downarrow$  & 21.8  $\uparrow$ & 47.8 $\downarrow$  \\
         \multicolumn{1}{r}{\textit{+self-attention}} & \textbf{72.5} $\uparrow$ & 21.2 $\downarrow$ & 60.4  $\downarrow$ \\
         \midrule
         Parsing (BiLRNN)               & 68.6  & 22.2  & 72.9  \\
         \multicolumn{1}{r}{\textit{+max-pooling}}       & 69.7 $\uparrow$ & 21.8  $\downarrow$     & 48.3 $\downarrow$  \\
         \multicolumn{1}{r}{\textit{+mean-pooling}}      & 58.0 $\downarrow$   & 21.2 $\downarrow$  & 50.7  $\downarrow$ \\
         \multicolumn{1}{r}{\textit{+self-attention}} & 72.2 $\uparrow$ & 21.5 $\downarrow$  & 69.1$\downarrow$  \\
         \cmidrule(lr){2-4}
         Balanced (BiLRNN)              & 69.6  & 22.3 & 76.0  \\
         \multicolumn{1}{r}{\textit{+max-pooling}}       & 70.6  $\uparrow$ & 21.6   $\downarrow$  & 48.5   $\downarrow$  \\
         \multicolumn{1}{r}{\textit{+mean-pooling}}      & 54.1   $\downarrow$  & 21.3   $\downarrow$  & 52.7 $\downarrow$    \\
         \multicolumn{1}{r}{\textit{+self-attention}} & \textbf{72.5} $\uparrow$ & 21.6 $\downarrow$   & 69.5 $\downarrow$   \\ 
         \cmidrule(lr){2-4}
		 Left (BiLRNN) & 67.7 & 21.6 & 72.9 \\
         \multicolumn{1}{r}{\textit{+max-pooling}}       & 71.2 $\uparrow$ &  20.5 $\downarrow$   &  47.6  $\downarrow$   \\
         \multicolumn{1}{r}{\textit{+mean-pooling}}      & 67.3 $\downarrow$   & 21.4 $\downarrow$   &  51.8 $\downarrow$   \\
		 \multicolumn{1}{r}{\textit{+self-attention}} & 72.1 $\uparrow$ & 21.6~--~& 70.2 $\downarrow$    \\
         \cmidrule(lr){2-4}
         Right (BiLRNN) & 68.7 & \textbf{23.1} & \textbf{80.4} \\
         \multicolumn{1}{r}{\textit{+max-pooling}}       & 71.6 $\uparrow$ &  21.6 $\downarrow$   &   48.4 $\downarrow$   \\
         \multicolumn{1}{r}{\textit{+mean-pooling}}      & 67.2 $\downarrow$   &  22.1 $\downarrow$   &   53.9 $\downarrow$   \\
         \multicolumn{1}{r}{\textit{+self-attention}}  & 72.4 $\uparrow$ & 21.6 $\downarrow$   & 68.9 $\downarrow$   \\
         \bottomrule
    \end{tabular} 
    \caption{\label{table:pooling} Performance of tree and linear-structured encoders with or without pooling, on the selected three tasks. We report accuracy $(\times 100)$, char-level BLEU for MT and word-level BLEU for AE. All of the tree models have bidirectional leaf RNNs (BiLRNN). The best number(s) for each task are in bold. The top and down arrows indicate the increment or decrement of each pooling mechanism, against the baseline of pure tree based encoder with the same structure.}
\end{table}

\section{Discussions}
\label{sec:related}

Balanced tree for sentence modeling has been explored by \newcite{munkhdalai2017neural} and \newcite{williams2018latent} in natural language inference~(NLI).
However, \newcite{munkhdalai2017neural} focus on designing inter-attention on trees, instead of comparing balanced tree with other linguistic trees in the same setting.
\newcite{williams2018latent} do compare balanced trees with latent trees, but balanced tree does not outperform the latent one in their experiments, which is consistent with ours.
We analyze it in Section \ref{sec:main-results} that sentences in NLI are too short for the balanced tree to show the advantage.

\newcite{P18-2116} argue that LSTM works for the gates’ ability to compute an element-wise weighted sum. In such case, tree LSTM can also be regarded as a special case of attention, especially for the balanced-tree modeling, which also automatically select the crucial information from all word representation.
\newcite{kim2017structured} propose a tree structured attention networks, which combine the benefits of tree modeling and attention, and the tree structures in their model are also learned instead of the syntax trees. 

Although binary parsing trees do not produce better numbers than trivial trees on many downstream tasks, it is still worth noting that we are not claiming the useless of parsing trees, which are intuitively reasonable for human language understanding. 
A recent work \cite{blevins2018deep} shows that RNN sentence encodings directly learned from downstream tasks can capture implicit syntax information.
Their interesting result may explain why explicit syntactic guidance does not work for tree LSTMs.
In summary, we still believe in the potential of linguistic features to improve neural sentence modeling, and we hope our investigation could give some sense to afterwards hypothetical exploring of designing  more effective tree-based encoders.

\section{Conclusions}
\label{sec:conclusions}
In this work, we propose to empirically investigate what contributes mostly in the tree-based neural sentence encoding. We find that trivial trees without syntax surprisingly give better results, compared to the syntax tree and the latent tree. Further analysis indicates that the balanced tree gains from its  shallow and balance properties compared to other trees, and right-branching tree benefits from its strong structural prior under the setting of left-to-right decoder.

\section*{Acknowledgements}
We thank Hang Li, Yue Zhang, Lili Mou and \mbox{Jiayuan} Mao for their helpful comments on this work, and the anonymous reviewers for their valuable feedback. 

\newpage
\bibliography{emnlp2018}
\bibliographystyle{acl_natbib_nourl}

\newpage
\appendix
\section{Implementation Details}
Our codebase is built on PyTorch 0.3.0.\footnote{\url{https://pytorch.org/docs/0.3.0}} All the sentences was tokenized with SpaCy.\footnote{https://spacy.io}

\subsection{Sentence Encoding}
We use LSTM based sentence encodings as the extracted features of sentences for downstream classification or generation tasks. We use typical long short term memory \cite[LSTM;][]{hochreiter1997long} units for linear structures, which can be summarized as:
\begin{equation*}
\begin{aligned}
\bm{f}_t & = \sigma(\bm{W}_f\cdot[\bm{h}_{t-1}, \bm{x}_t] + \bm{b}_f) \\
\bm{i}_t & = \sigma(\bm{W}_i\cdot[\bm{h}_{t-1}, \bm{x}_t] + \bm{b}_i) \\
\tilde{\bm{c}}_t & = \tanh(\bm{W}_c\cdot[\hm{h}_{t-1}, \bm{x}_t] + \bm{b}_c) \\
\bm{o}_t & = \sigma(\bm{W}_o\cdot[\bm{h}_{t-1}, \bm{x}_t] + \bm{b}_o) \\
\bm{c}_t & = \bm{f}_t \bm{c}_{t-1} + \bm{i}_t\tilde{\bm{c}}_{t} \\
\bm{h}_t & = \bm{o}_t \tanh(\bm{c}_t)
\end{aligned}
\end{equation*}
where $t$ indicates the time step of a state; $\bm h_t$ is the hidden state and $\bm x_t$ is the input vector. 
We apply binary tree LSTM units adapted from \newcite{zhu2015long} for binary tree LSTMs, which can be summarized as:
\begin{equation*}
\begin{aligned}
\bm{f}_l & = \sigma(\bm{W}_l\cdot[\bm{h}_{l}, \bm{h}_r] + \bm{b}_l) \\
\bm{f}_r & = \sigma(\bm{W}_r\cdot[\bm{h}_{l}, \bm{h}_r] + \bm{b}_r) \\
\bm{i}_t & = \sigma(\bm{W}_i\cdot[\bm{h}_{l}, \bm{h}_r] + \bm{b}_i) \\
\tilde{\bm{c}}_t & = \tanh(\bm{W}_c\cdot[\hm{h}_{l}, \bm{h}_r] + \bm{b}_c) \\
\bm{o}_t & = \sigma(\bm{W}_o\cdot[\bm{h}_{t-1}, \bm{x}_t] + \bm{b}_o) \\
\bm{c}_t & = \bm{f}_l \bm{c}_{l} + \bm{f}_r \bm{c}_r + \bm{i}_t\tilde{\bm{c}}_{t} \\
\bm{h}_t & = \bm{o}_t \tanh(\bm{c}_t)
\end{aligned}
\end{equation*}
where the subscript $t$ denotes the current state, and $l, r$ denote the left and right child states respectively. We also apply LSTM \cite{hochreiter1997long} as leaf-node RNN when necessary. 

It is worth noting that left-branching tree LSTM without leaf-node RNN is structurally equivalent to unidirectional LSTM. The only difference between them, which may cause the slight difference on performance, comes from the implementation of LSTM units. 

The candidate set of dropout ratio we explore for the task of word-level semantic relation (WSR) is $\{0, 0.1, 0.15, 0.2, 0.3, 0.5\}$.

\subsection{Sentence Relation Classification}
In the task of sentence relation classification, the feature vector consists of the concatenation of two sentence vectors, their difference, and their element-wise product \cite{mou2016natural}:
\begin{equation*}
\centering
\bm{z} = \left(
\begin{aligned} 
&\bm{s}_1 \\
&\bm{s}_2 \\
\bm{s}_1 &- \bm{s}_2 \\
\bm{s}_1 &\odot \bm{s}_2
\end{aligned}
\right)
\end{equation*}

\subsection{Pooling Mechanism}
Following \cite{socher2011dynamic}, we apply pooling mechanism to all leaf states (of tree LSTMs) and hidden states.  The detailed pooling methods are described as follows. 

\paragraph{Max Pooling.}
Max pooling takes the max value for each dimension 
\begin{equation*}
\begin{aligned}
\bm{H} &= (\bm{h}_1, \bm{h}_2, \cdots, \bm{h}_{m}) \\
\bm{s}_i &= \max_{j=1}^m h_{j,i} ~~~~~  i=1, 2, \cdots, d \\
\bm{s} &= (\bm{s}_1, \bm{s}_2, \cdots, \bm{s}_d)^T
\end{aligned}
\end{equation*}
where $\bm{h}_i$ denotes a leaf state in tree LSTMs or a hidden state; $m=2n-1$ for tree LSTMs and $m=n$ for linear LSTMs; $\bm{s}$ denotes the final sentence encoding.
\paragraph{Mean Pooling.}
Mean pooling (average pooling) takes the average of all hidden states as the sentence representation, which can be summarized as: 
\begin{equation*}
\begin{aligned}
\bm{H} &= (\bm{h}_1, \bm{h}_2, \cdots, \bm{h}_{m}) \\
\bm{s} &= \frac{1}{m} \sum_{i=1}^m \hm{h}_i
\end{aligned}
\end{equation*}

\paragraph{Self-Attention.}
We follow \newcite{conneau2017supervised} and \newcite{lin2017astructured} to build a self-attentive mechanism, which can be summarized as:
\begin{equation*}
\begin{aligned}
\bm{H} & = (\bm{h}_1, \bm{h}_2, \cdots, \bm{h}_{m}) \\
\bm{a} & = \mathrm{softmax}(\mathbf{w}_\beta^T \tanh (\mathbf{W}_\alpha \bm{H})) \\ 
\bm{s} & = \bm{H} \bm{a}^T
\end{aligned}
\end{equation*}
where $\bm{a}$ denotes attention weights computed by learned parameters $\mathbf{W}_\alpha$ and $\mathbf{w}_\beta$.
In all experiments, $\mathbf{w}_\beta$ is a 128-d vector.

\end{document}